\documentclass[a4paper, 10pt, conference]{ieeeconf}      

\IEEEoverridecommandlockouts                              

\usepackage[utf8]{inputenc}
\usepackage[english]{babel}

\usepackage{todonotes}

\usepackage{cite}
\usepackage{multirow}
\usepackage{graphicx} 
\usepackage{epsfig} 
\usepackage{booktabs}

\usepackage{amsmath} 
\usepackage{amssymb}  

\usepackage{flafter} 

\usepackage{caption}
\usepackage{subcaption}

\usepackage{algorithm}
\usepackage[]{algpseudocode} 

\usepackage{program}

\usepackage{breqn}

\usepackage{siunitx} 
\usepackage{tabularx}

\usepackage{hyperref}
\usepackage{cleveref}

\title{\LARGE \bf
Preparation of Papers for IEEE Sponsored Conferences \& Symposia*
}

\graphicspath{{figures/}}

\title{\LARGE \bf Creation of a Modular Soft Robotic Fish Testing Platform}%

\author{Yu Zhang$^{1}$, Robert K. Katzschmann$^{1}$ 
\thanks{$^{1}$ Soft Robotics Lab, ETH Zürich, Switzerland.}
\thanks{{\tt\footnotesize \href{mailto:zhangyu@ethz.ch}{zhangyu@ethz.ch}, \href{mailto:rkk@ethz.ch}{rkk@ethz.ch}}}
}

\begin{document}

\maketitle
\thispagestyle{empty}
\pagestyle{empty}


\begin{abstract}
Research on the co-optimization of soft robotic design and control requires rapid means for real-world validation. Existing creation pipelines do not allow for the swift prototyping of soft robots to quickly test various design configurations and control policies.
This work proposes a pipeline for rapid iterative design and fabrication of a miniaturized modular silicone-elastomer-based robotic fish. The modular design allows simple and rapid iterations of robotic fishes with varying configurations to assist current research efforts on the development of design optimization methods. The proposed robotic fish can serve as a standardized test platform on which performance metrics such as thrust and range of motion can be evaluated. We further show the design of an underwater evaluation setup capable of measuring input pressure, tail deformation, and thrust. Multiple robotic fish prototypes with varying stiffness and internal pneumatic chamber configurations are fabricated and experimentally evaluated. The presented flexible modular design principle for the robot and its evaluation platform unlocks the possibilities of more efficient soft robotic fish and will benefit research on design optimization and underwater exploration in the future.







\end{abstract}


\section{Introduction} \label{sec:introduction}

\subsection{Motivation}
Soft robots, unlike traditional robots with rigid links and actuators, use elastic and compliant materials in their construction and actuation. The inherent compliance of soft robots allows for a safer integration into the daily lives of humans -- a key breakthrough in robotic technology development. The soft and compliant nature of such robots is inspired by real muscles and other soft tissues used as nature's building blocks in the creation of living beings. The similarities between soft robots and living beings not only eases the realization of bio-mimetic robotic designs, but also deepens the engineering understanding of nature's optimized creations.

Fish are mesmerizing animals that have been optimized throughout evolution to achieve excellent underwater performance in various natural scenarios and tasks. The ability to understand and create life-like robotic fishes has been pursued for a long time by the robotics community. Despite the promise and potential shown by soft robots, their compliant nature also brings forward new challenges when creating their complex composition, modeling their high degrees of freedom, and controlling them as underactuated dynamical systems\cite{rus_design_2015}.

\begin{figure}[t]
        \centering
        \includegraphics[width=\columnwidth]{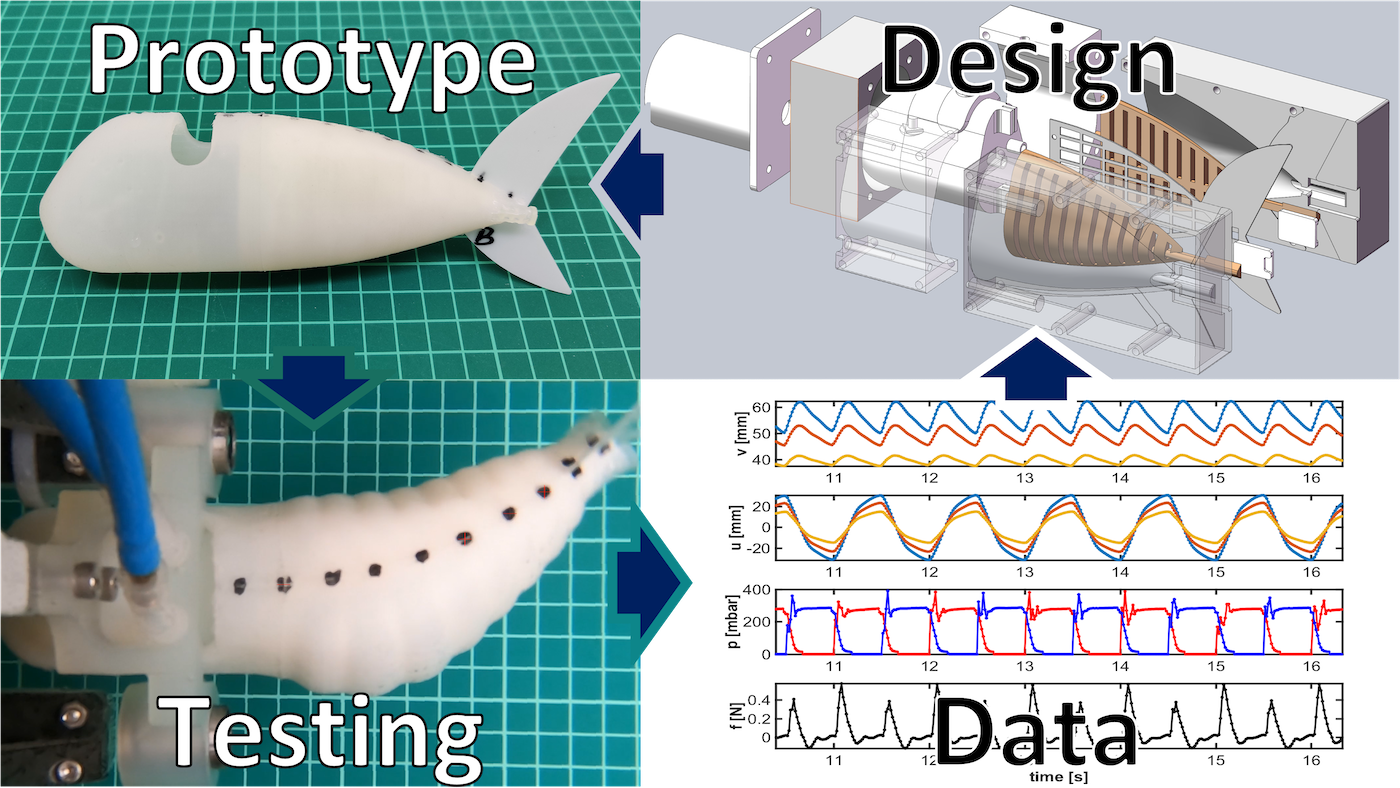}
            \caption{The modular robotic fish design and the evaluation platform enable an iterative design optimization process. Designs generated from an optimization algorithm can be rapidly fabricated and evaluated for performance and the resulting data can be fed back into an optimization algorithm to generate the next design iteration.}
        \label{fig:overview}
\end{figure}


\subsection{Prior Art and Challenges}
\label{sec:related}
A range of optimization methods, including Deep Learning-based\cite{sasaki_topology_2019} and Finite-Element Modeling (FEM)-based optimization\cite{ma_diffaqua_2021}, have been developed in an attempt to represent and optimize the design and construction of soft robotic fishes. The iterative nature of such optimization pipelines requires that multiple iterations of robots have to be fabricated and tested. However, current methods to fabricate robotic fishes\cite{schaffner_3d_2018,wallin_3d_2018} remain highly tailored for each specific design~\cite{katzschmann_exploration_2018,zhu_tuna_2019,berlinger_implicit_2021,li_self-powered_2021}. This lack of modularity and versatility means that a dedicated set of tools and components will need to be fabricated for each design iteration~\cite{cho_review_2009}. 
This approach of realization significantly slows down research progression while also creating hardware barriers for algorithmic developments.

\subsection{Approach}
The lack of accessible functional robotic hardware (see \Cref{sec:related}) motivates the development of a modular robotic fish platform for rapid creation and evaluation. Featuring modular molds for different sections of the fish, a family of robots with different size, shape, and actuation can be fabricated by simply using different combinations of molds. The modular mold-set, when assembled, allows for a single cast based on the lost-wax principle.
The lost-wax-casting fabrication method for soft robots~\cite{katzschmann_hydraulic_2016} was originally conceived to avoid multiple tedious step of additive casting by requiring only a single casting step to form an entire soft robotic body. This uni-body forming technique also eliminates the problem of reliably interfacing soft and stiff sections of a soft robot while substantially reducing time and skill required during fabrication. 
To evaluate the performance of a created robot, an experimental setup capable of simultaneously measuring pressure input, tail deflection, and thrust is also developed. An overview of the enabled design optimization pipeline with the proposed robotic fish platform is provided in \Cref{fig:overview}.

\subsection{Significance}
Apart from serving as a test bench for co-optimization, the robot also has the future potential to be fitted with different soft actuators such as DEAs\footnote{Dielectric Elastomer Actuators}\cite{ohalloran_review_2008} and HASELs\footnote{Hydraulically Amplified Self-healing ELectrostatic actuators}~\cite{acome_hydraulically_2018}. The test bench essentially allows for a comparative performance study against traditional PneuNet\footnote{Pneumatic Networks}\cite{marchese_recipe_2015} actuators. With a small form factor and simple fabrication procedure, the robotic fishes can also be mass-produced and used as the hardware platform in studies on the swarming and schooling behavior of underwater swimmers. 

\subsection{Contributions}
In summary, we contribute with this work:
\begin{itemize}
    \item the design and fabrication of a robotic fish platform facilitating the design optimization of soft robotic fish;
    \item a highly modular mold-casting pipeline to rapidly fabricate fluidically-actuated soft robotic fish using soft silicone elastomers and flexible backbones; and
    \item an experimental evaluation of multiple robot designs on the above platform.
\end{itemize}

\subsection{Paper Outline}
The design and fabrication process of the platform are detailed in \Cref{sec:design} and \Cref{sec:fabrication}. The creation of the platform is then followed by a detailed experimental evaluation \Cref{sec:evaluation}. The measurements the platform can record is then shown in \Cref{sec:results}. A conclusion and discussion is then provided in \Cref{sec:conclusion}.

\section{Design}
\label{sec:design}
This section documents the key methods and considerations in the design process for a silicone-elastomer-based soft robotic fish. The design of the molds and cores for the later fabrication is also described.
    
    \subsection{Design of the miniaturized robotic fish} 
        The design of the robotic fish is based on the identified key functional requirements. The length of the robotic fish is set to be under \SI{0.2}{\meter} to reduce cost and fabrication complexity compared to previous designs that are around \SI{0.47}{\meter} long~\cite{katzschmann_exploration_2018}. Despite the aim of having a short robotic fish that can be formed in one casting step, the ability to quickly fabricate a different prototype is achieved by splitting the fish into three sections with separate molds: the tail section, the body section, and the head section.
        
        The main actuation of the robotic fish is set to be a hydraulic tail actuator. Inspired by \emph{SoFi}~\cite{katzschmann_hydraulic_2016}, the actuator consists of two "ribbed" hydraulic chambers situated on both sides of an embedded flexible fin at the center of the fish. The center fin serves as a structural constraint and provides thrust during the undulating motion induced by the alternating expansion of the two hydraulic chambers under pressure. The ratio between pneumatic chamber thickness and wall thickness is set to 3:1 based on preliminary tests and the wall thickness is set to \SI{2}{\mm}. The shape of the hydraulic chambers are defined according to the external shape of the tail with a \SI{3}{\mm} thick skin around the internal chambers.
        
        Once the tail section is designed, the body section is constructed by extending the cross section at the front of the tail actuator forward. The body section is then hollowed out with a skin thickness of \SI{3}{\mm}. This operation forms a large internal cavity for housing any driving electronics or motors of the design. A slotted cut out is added at the top of the body section to allow access to the internal electronics. The slotted design also reduces the chance of tearing the robotic skin during the mold extraction after casting.
        
\begin{figure}[htbp]
        \centering
        \includegraphics[width=\columnwidth]{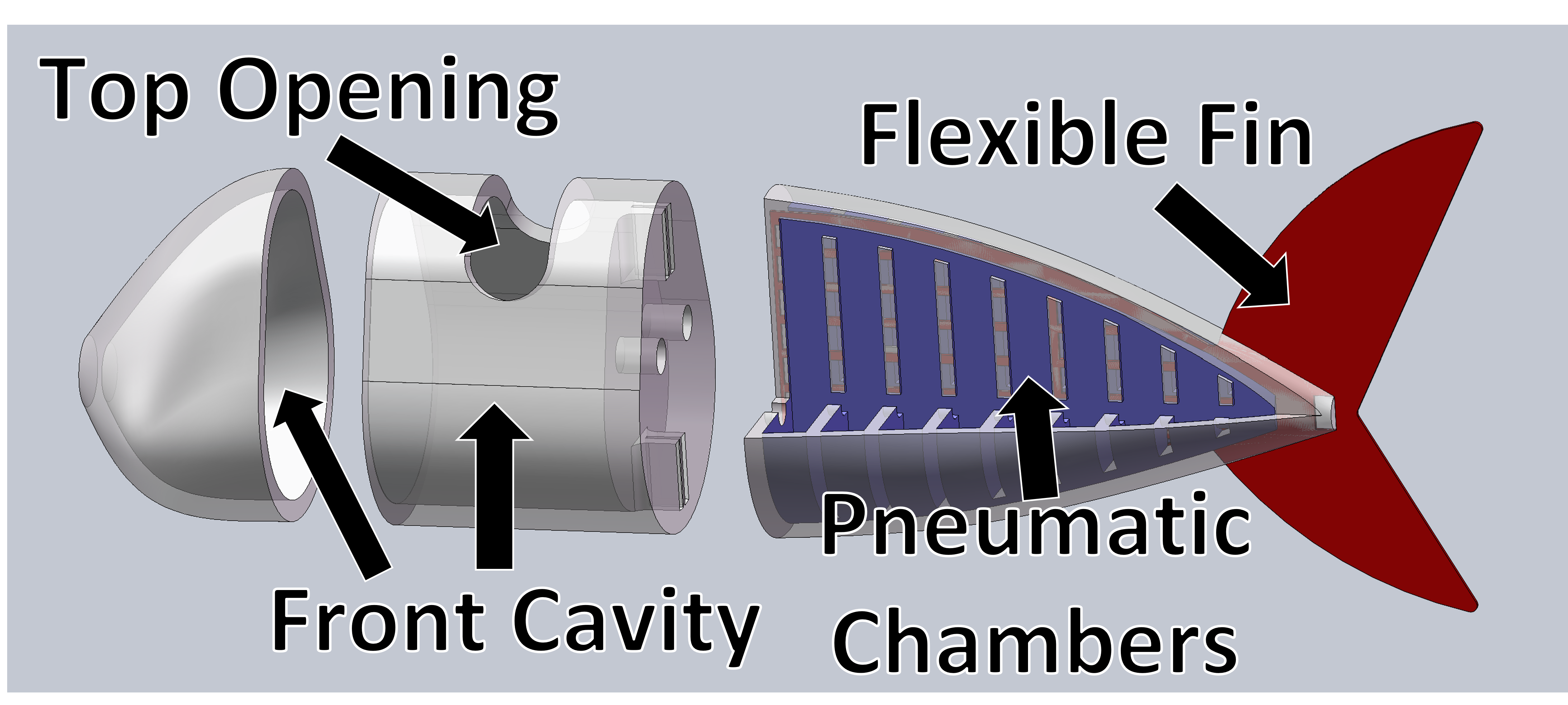}
            \caption{An exploded view of the robotic fish design. All the separate sections are merged and formed in one cast into a final robot. The internal chamber walls are colored blue for better visualization. The flexible central spine is colored red.}
        \label{fig:exploded view}
        \end{figure}

        Finally, the head of the robot is designed to be 30 millimeters long with a simple round shape at the front. This section is again hollowed out with a skin of \SI{3}{\mm} thickness. The cavity in the head section connects to the cavity in the body section to form one bigger cavity for internal electronics. Being at the front of the robot, the head cavity can be fitted with sensors such as SONAR\footnote{Sound Navigation and Ranging} sensors~\cite{shin_fish_2007} and cameras~\cite{katzschmann_exploration_2018} to explore the surrounding environment and exchange information\cite{berlinger_implicit_2021} with other robots in a study on swarming.
        An exploded view of the robotic design is shown in \Cref{fig:exploded view}.
        
    \subsection{Design of corresponding cores and molds}
        Given the robotic fish shape including the internal chamber structure and the external shape,  corresponding mold pieces are developed. Since the internal hydraulic chambers have a complex ribbed structure, traditional molds with 3D printed plastic cannot be used due to great difficulties in mold extraction. Cores made out of wax can be melted away and removed easily. This insight leads to the adoption of lost-wax casting\cite{katzschmann_hydraulic_2016} in the fabrication process. To fabricate the robotic fish, three types of molds have to be designed: internal sacrificial wax cores, removable internal cores, and external molds. 
        
        The internal wax cores are used in the final mold assembly to create the pneumatic chambers, the shape of which can be extracted from the robot design by performing an intersection operation between the robotic fish and an arbitrary block shape. Support structures such as alignment notches used to fix the location of the cores in the mold assembly are also added. The wax cores, as they cannot be directly printed, require a set of soft negative silicone molds with corresponding negative plastic molds. Such plastic molds, after being negated twice, resemble the same shape as the targeted wax cores. After the core shape is isolated, a bottom plate and surrounding walls are added to form the plastic mold design for the silicone mold.
        
        Similar to the internal wax cores, the removable internal core shape is also extracted from the robot design. Additional interfacing structures are then added to finish the internal core design. These interfacing structures include mounting holes for bolts during mold assembly, alignment structures used to prevent core dislocation during casting, and clearance cutouts used to house the central spine and attach to the wax cores from the tail section.
        
        The external molds are separated into tail, body, and head sections to allow for modular exchange between various designs. The design process for the external molds is similar to that of the wax core molds. After the external shape is isolated with an intersection operation, extra cut-outs are made to host the central spine and the alignment notches on the wax cores. Finally, the opposite mold piece is mirrored and fastener holes are cut out. To assist in the casting process, a funnel piece is also designed around the outer dimensions of the mold and then added on top during the mold assembly. 
        
\begin{figure*}[htbp]
        \centering
        \includegraphics[width=\textwidth]{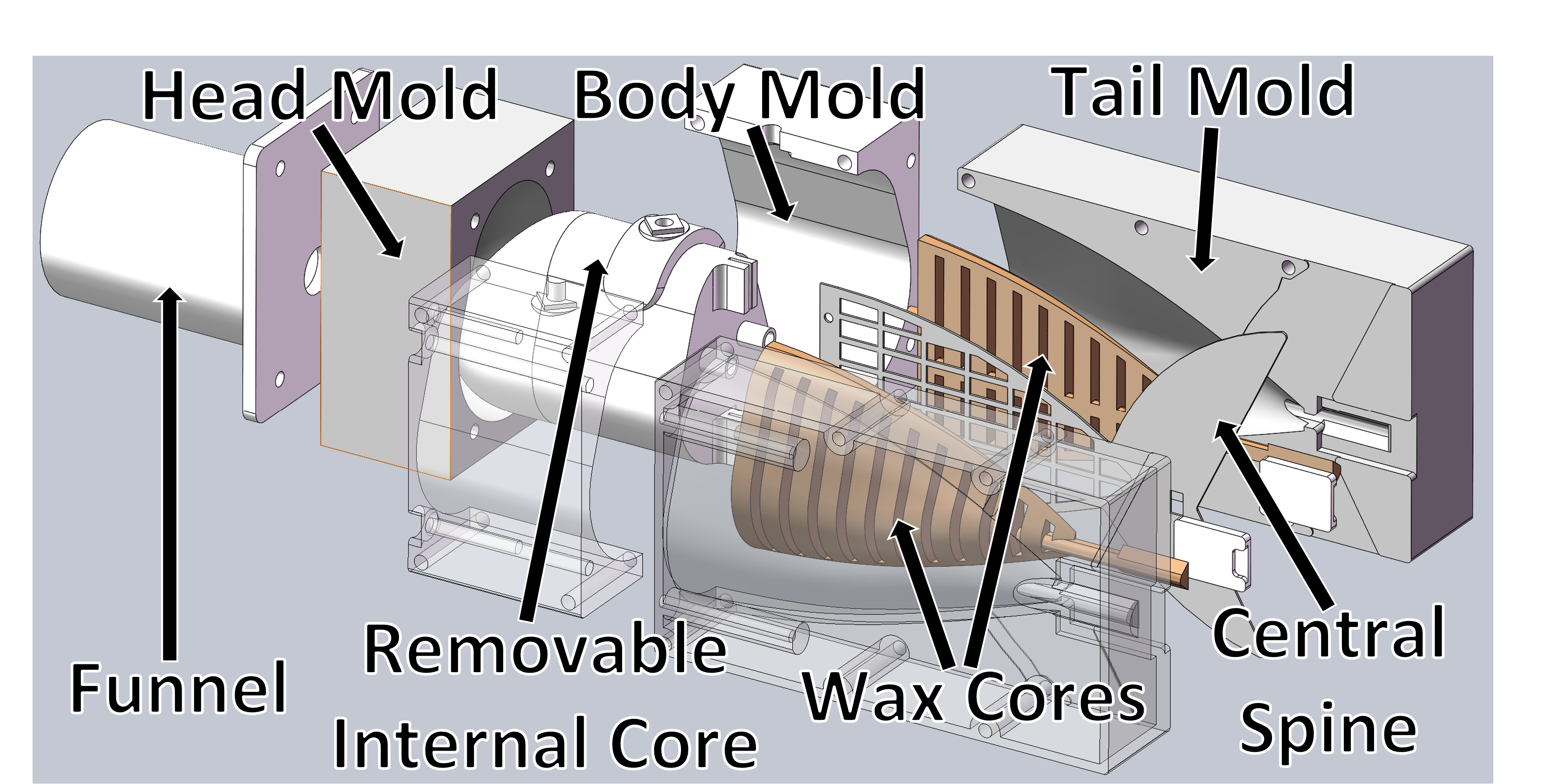}
        \caption{Exploded view of the modular mold assembly. Note that fasteners are omitted for clarity. The two wax cores are fixed to the tail molds before the central spine is fixed in between the tail molds as fasteners clamp the tail molds together. The removable internal core is then placed next to the wax cores and attached to the central spine and wax cores via a slot and two extrusions on the back. The body molds can then be installed from the side and fastened in place before a bolt attaches the internal core to the body molds. This bolt then fully defines the position of the internal core, wax cores, and the central spine. Finally, the head mold and funnel are attached and fastened to form the final mold assembly.}
        \label{fig:molds_design}
        \end{figure*}

        An exploded view of the finished mold assembly is shown in \Cref{fig:molds_design}. The design is generated in SolidWorks 2020 by Dassault Systèmes.
        
\section{Fabrication}
\label{sec:fabrication}
To cast the robotic fish, the wax cores must be fabricated first. After molds for the silicone mold are 3D printed and prepared by cleaning and smoothing the surface with a knife and sand paper, a thin layer of mold release is sprayed on to assist in the de-molding process later. Silicone elastomer\footnote{Smooth-On Dragon Skin 20 2-Part Silicone with shore hardness \SI{20}{A}} is mixed, degassed in vacuum, and then poured into the mold. The molds are then put into vacuum again to further degas the silicone and allow it to cure after the vacuum is lifted later. The extracted silicone molds are then put into the oven with beeswax to prepare for the wax casting. Two \SI{2.5}{\mm} carbon-fiber rods, serving as structural support, are cut to length and then added into the corresponding slots in the silicone mold to heat up as well. Melted wax is poured into the hot silicone mold and observed during the cooling down process; if a shrinkage of the cooling wax occurs, extra wax will be added. After the wax has fully solidified, a razor blade is used to scrape the top surface of the silicone mold to remove any excess wax. The deformable silicone mold allows for a simple extraction of the cores, where the molds are slightly bent such that the finished cores can be carefully separated.

After laser cutting the central spine out of \SI{0.5}{\milli\metre} polyoxymethylene sheets and spraying all mold pieces with mold release, the final mold set can be assembled. The alignment structures on the wax cores are fitted into the corresponding slots before a piece of plastic spacer is pushed in to secure the wax cores. The central spine is placed in the middle of the two halves of the tail mold and fitted into the slots on the two mold pieces. The tail mold module can then be closed and fastened with bolts.

To form the body and head section, the removable internal core is placed on top of the tail mold module. The two holes on the bottom of the internal core fit to the two extruding rods on the wax cores while the central spine fits to a slot. The two halves of the body molds are then installed from the side and fastened with bolts running horizontally. Four vertically running bolts are then used to attach the body mold module to the tail module. With the body molds fixed, the internal mold is attached, through alignment structures, to the body mold module. This attachment ensures that the internal mold piece is also rigidly fixed and will not move during the casting process. What remains to be installed are the head module and the funnel to house the expanding silicone during degassing, those are simply stacked and mounted with four vertically running bolts.

Similar to the wax core mold, silicone elastomer is poured into the mold assembly after mixing and degassing. However, the casting process of the robotic fish slightly differs from the casting of the wax core molds due to more complex internal structures. This difference effectively means that it will be harder for internal air bubbles to be removed and a new casting process with multiple degassing stages is adopted. The mold assembly is placed on a specially designed frame to tilt the mold assembly to one side before pouring in the silicone. This tilting position makes it easier for the air bubbles to rise up out of the mold from the thin gaps in the wax cores. After the first round of degassing in vacuum, the mold assembly is flipped around on the frame and degassed again to remove air bubbles from the other side of the wax cores. Finally, the mold assembly is placed straight and allowed to cure.

\begin{figure}[htbp]
    \centering
    \includegraphics[width=1\columnwidth]{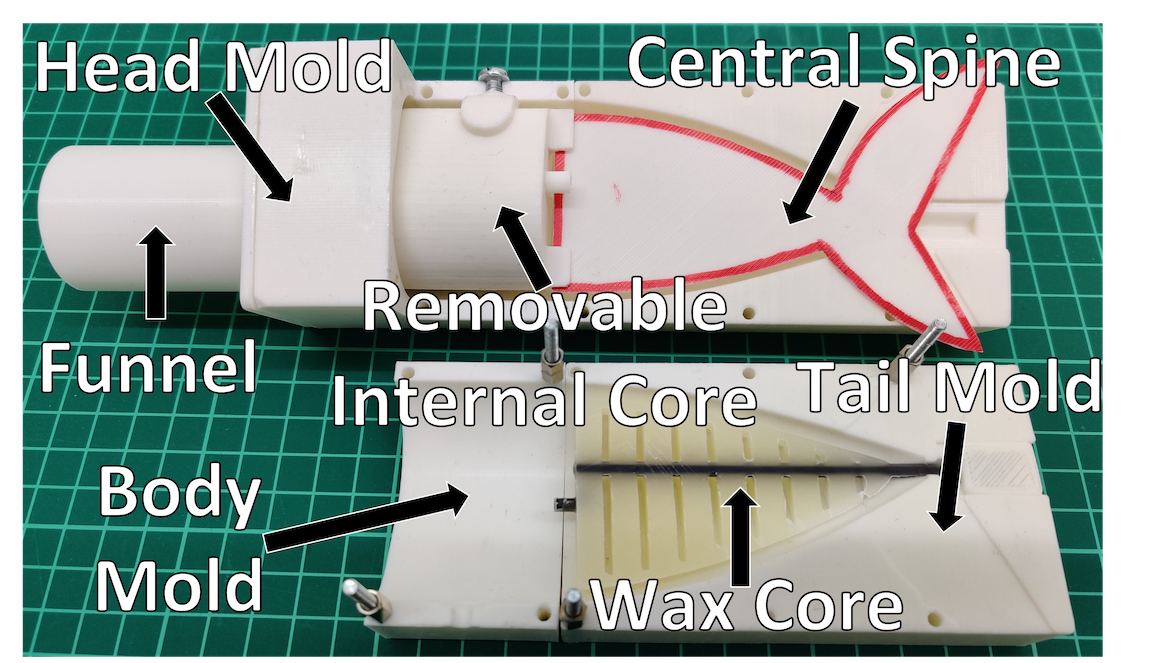}
    \caption{Partially assembled mold assembly used in the fabrication process. A grid pad with \SI{10}{\mm} grid is placed in the back ground. The central spine is replaced with a painted sheet for demonstration purposes. During casting, the silicone elastomer flows around the gaps between the molds and cores to form the skin and internal chambers of the soft robotic fish.}
    \label{fig:molds}
\end{figure}

\begin{figure}[htbp]
    \centering
    \includegraphics[width=1\columnwidth]{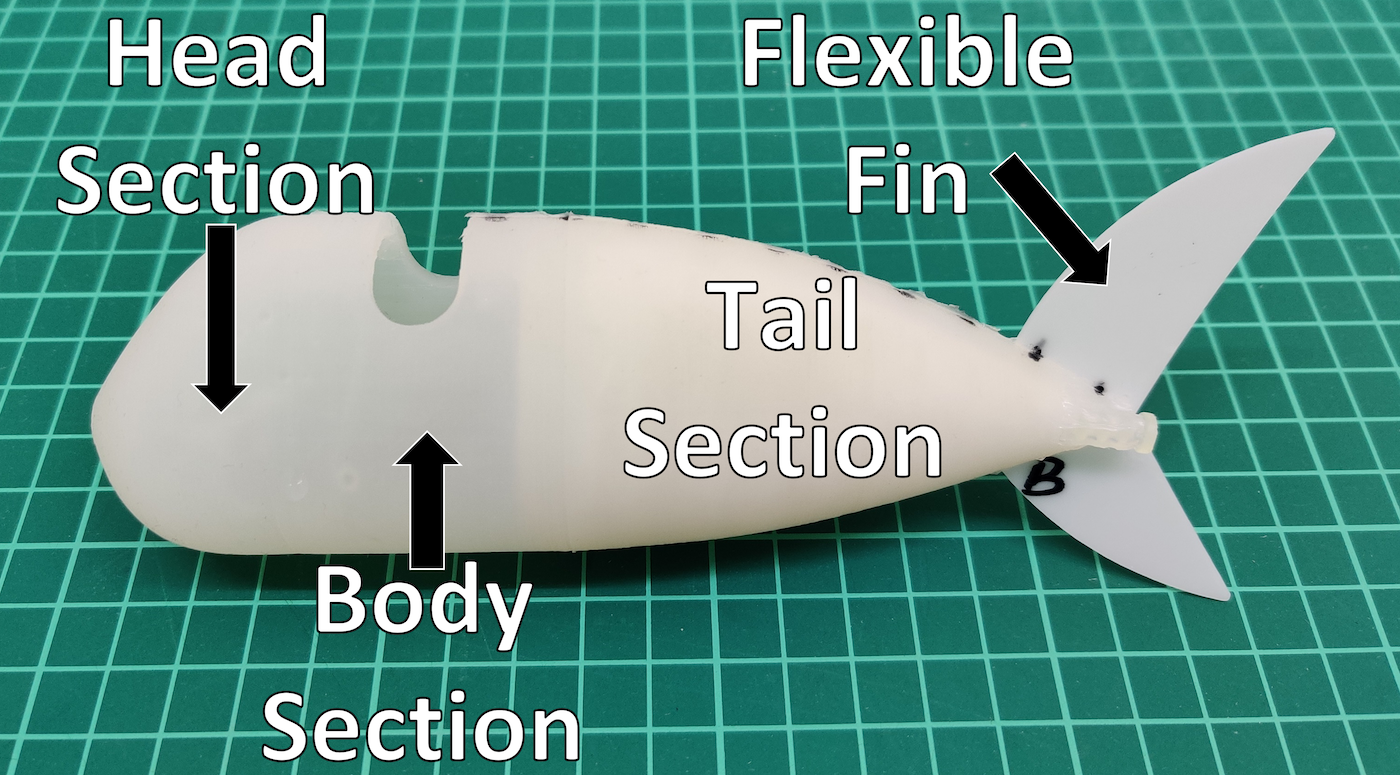}
    \caption{Labeled photo of Bruce, a robotic fish made with Smooth-On DragonSkin 20 silicone elastomer with a shore hardness of \SI{20}{A}. A grid pad with a \SI{10}{\mm} grid is placed in the background.}
    \label{fig:fish_final}
\end{figure}

The disassembly of the mold after curing must follow a certain sequence described below to avoid damaging the finished robotic fish. The four vertical bolts holding the head mold and the funnel must be removed first before a thin blade slices through the space between the funnel and the head mold to trim away the sacrificial silicone from the robotic fish. This operation frees the funnel, which can now be removed. The horizontal bolt holding the internal mold must then be removed to free up the internal mold and avoid tearing the silicone skin during later extraction steps. The head mold and body molds can then be disassembled and removed. Finally, the tail mold must be opened from the front side. This is because that the wax cores are still fixed with the two tail molds at the end and opening the tail mold there can risk putting excess force on the wax cores and tearing the tail. After the tail molds are removed, the final step will be to carefully pull on the front skin to separate the internal mold before the skin is carefully peeled back to extract the internal mold. After trimming away any excess silicone, the robotic fish is placed head-up in the oven to melt out the wax cores. The fabrication process of the final robotic fish will be finished after two silicone tubes are glued into the tail end and plugged with a pair of 3D printed plugs. The mold assembly used in the fabrication process is shown in \Cref{fig:molds} and one of the finished robotic fish is shown in ~\Cref{fig:fish_final}.

\section{Evaluation}
\label{sec:evaluation}


An underwater experiment setup is constructed to evaluate the performance of the finished robots. A rubber pad with \SI{10}{\mm} grid lines is placed at the bottom of the tank to provide visual reference for the deformation measurements. A camera\footnote{GoPro Hero 6 Black} is mounted on top of the tank with a bird's eye view of the setup and the bottom grid pad. The camera is also fixed in a way that the camera lens iss submerged under water to minimize distortions at the water-air interface. A line of black dots with a horizontal distance of 10 millimeters were painted on the back of the fish robot to be later tracked by the camera~\cite{lucas1981iterative,tomasi1991tracking}.
Similar to the removable internal core during casting, the robots are attached to a mounting piece with the shape of the front cavity. The mounting piece is then mounted to a load cell\footnote{TAL2210, range \SI{1}{\kg}} to measure thrust forces during actuation. The thrust data is then processed by an amplifier\footnote{HX711 set to 80Hz data rate}, time-stamped, and transmitted with a micro-controller\footnote{Arduino Leonardo} via serial to the logging computer. Two \SI{8}{\mm} linear bearings with fitting smooth rods are initially installed to reduce the effects of side-ways forces on the load cell and improve the accuracy of the thrust measurement. However, the rails are removed during the final experiments since the friction has a larger effect on the thrust measurement.

The two pneumatic chambers of the robotic fish are connected to two output channels from a Festo proportional valve manifold\footnote{MBA-FB-VI, \SI{0}-\SI{2}{bar} output range with 1\% accuracy}. The valve manifold generates a internally regulated pressure output according to input commands and provides time-stamped log data for both input commands and measured output pressure at a frequency of \SI{200}{\Hz}.

\begin{figure}[htbp]
            \centering
            \includegraphics[width=\columnwidth]{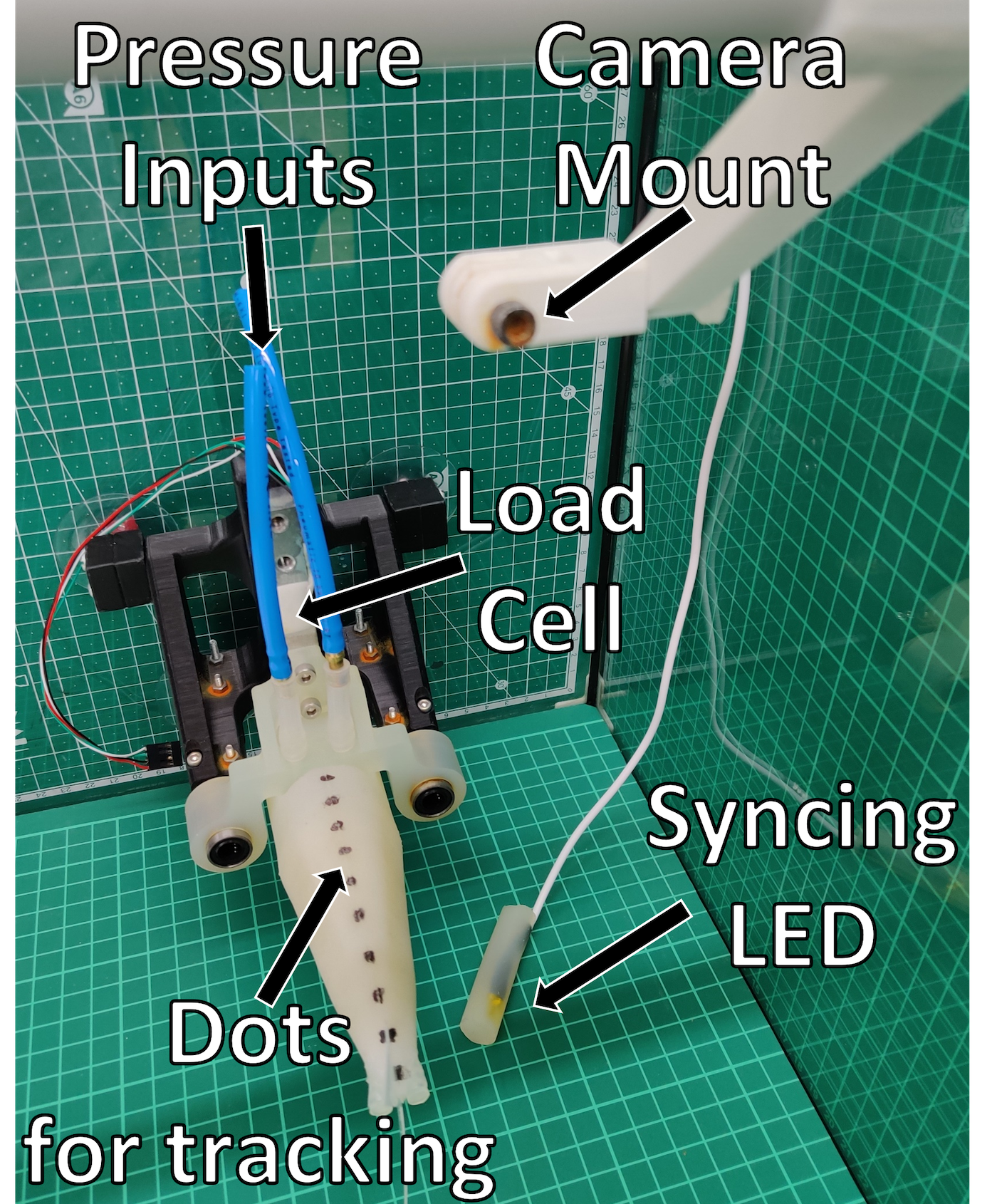}
            \caption{Labeled photo of the test setup. The front of the robot was wrapped around a mounting piece the shape of the front cavity. The mounting piece was fixed via the load cell to the experiment test jig. The test jig was attached to the side of the water tank with 4 suction cups. Pneumatic pressure from the Festo valve manifold is connected to the mounting piece and then into the pneumatic chambers. A camera was placed on top of the experiment setup with a bird's eye view. An led was placed on the bottom of the tank to sync the data from different measurements. }
            \label{fig:test_setup}
\end{figure}

The force and pressure measurements are activated with a single command to ensure syncing of the measured data. To sync visual measurements from the camera, a waterproof LED is placed in frame of the camera and lights up at the start of the measurement to be captured in the video footage. The frame in which the LED lights up can then be identified and used to sync the deformation measurement.
A labeled photo of the experiment setup is provided in \Cref{fig:test_setup}.

Three different robotic fish are fabricated and evaluated with the experimental setup where pressure input, tail deformation, and thrust are measured. The three robotic fish are designed to have different material stiffness and internal pneumatic chamber configuration to demonstrate the adaptability of the fabrication process. Only the wax cores are changed for different chamber configurations, while only the silicone elastomer is changed to achieve different material stiffness. We also constructed a fourth robotic fish with DS 10 \footnote{Smooth-On Dragon Skin 10 2-Part Silicone with shore hardness \SI{10}{A}} and a thicker central fin\footnote{Laser-cut polyoxymethylene} after changing the tail mold section to a version with a larger fin cutout. However, the stiffness of the central fin prevents the robot from actuating and no further evaluations are done on this prototype. A Fifth robotic fish with DS 30 \footnote{Smooth-On Dragon Skin 30 2-Part Silicone with shore hardness \SI{30}{A}} was also constructed, but was not tested due great difficulties in the internal core removal process. The higher stiffness of the silicone elastomer prohibited the un-wrapping of the robotic fish off of the internal core.
For each robot, actuation signals in the form of pressure inputs with different amplitude and frequency are applied to demonstrate the potential of the robotic fish platform to be used in research on design and control optimizations.
The configurations of the three robot fish as well as the actuation signal inputs are shown in \Cref{tab:experiment_setup}.

\begin{table*}[htbp]

    \centering
    \caption{Robot configurations and actuation signals used for the experiments.}
    \resizebox{0.85\textwidth}{!}{%
        \begin{tabular}{@{}llll@{}}
        \toprule
        Robot name & \emph{Nemo} & \emph{Dory} & \emph{Bruce}\\ 
        \midrule
        Material & DS10 & DS10 & DS20\footnotemark \\ 
        Chambers & 9 & 12 & 9 \\
        Actuation signal type 1 & \SI{200}{mBar} at \SI{1}{Hz} & \SI{350}{mBar} at \SI{1}{Hz}& \SI{500}{mBar} at \SI{1}{Hz}\\
        Actuation signal type 2 & \SI{200}{mBar} at \SI{2}{Hz} & \SI{350}{mBar} at \SI{2}{Hz}& \SI{500}{mBar} at \SI{2}{Hz}\\
        Actuation signal type 3 & \SI{200}{mBar} at \SI{3}{Hz} & \SI{350}{mBar} at \SI{3}{Hz}& \SI{500}{mBar} at \SI{3}{Hz}\\
        Actuation signal type 4 & \SI{200}{mBar} at \SI{4}{Hz} & \SI{350}{mBar} at \SI{4}{Hz}& \SI{500}{mBar} at \SI{4}{Hz}\\
        Actuation signal type 5 & \SI{300}{mBar} at \SI{1}{Hz} & \SI{500}{mBar} at \SI{1}{Hz}& \SI{750}{mBar} at \SI{1}{Hz}\\
        Actuation signal type 6 & \SI{300}{mBar} at \SI{2}{Hz} & \SI{500}{mBar} at \SI{2}{Hz}& \SI{750}{mBar} at \SI{2}{Hz}\\
        Actuation signal type 7 & \SI{300}{mBar} at \SI{3}{Hz} & \SI{500}{mBar} at \SI{3}{Hz}& \SI{750}{mBar} at \SI{3}{Hz}\\
        Actuation signal type 8 & \SI{300}{mBar} at \SI{4}{Hz} & \SI{500}{mBar} at \SI{4}{Hz}& \SI{750}{mBar} at \SI{4}{Hz}\\
        \bottomrule
        \end{tabular}}
        \footnotetext{Smooth-On Dragon Skin 10/20 2-Part Silicone with shore hardness 10A/20A}
    \label{tab:experiment_setup}
\end{table*}

\begin{figure}[htbp]
            \centering
            \includegraphics[width=\columnwidth]{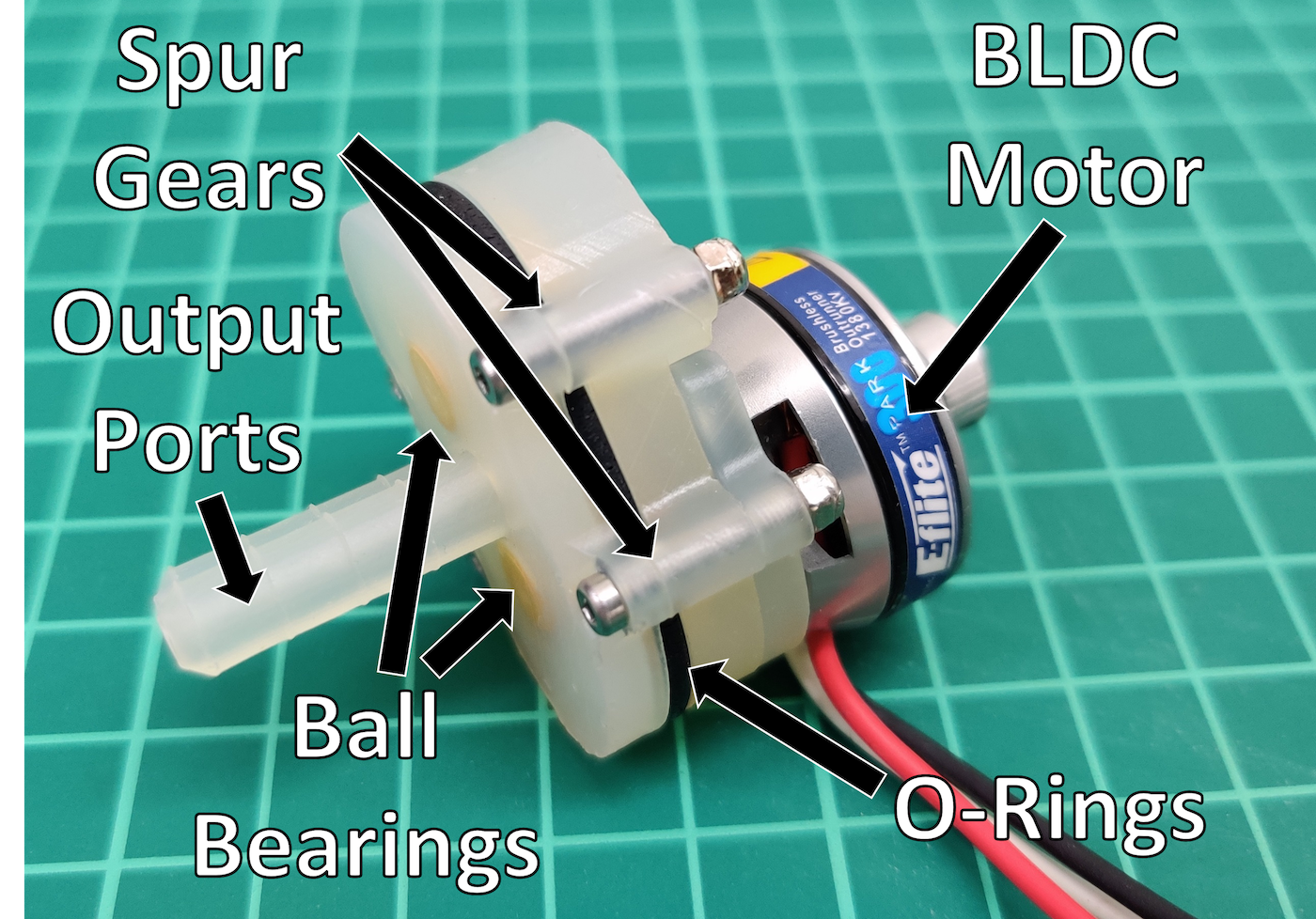}
            \caption{Labeled photo of the miniaturized gear pump developed to provide hydraulic power to the robotic fish. A grid pad with \SI{10}{mm} grid is placed in the back ground.}
            \label{fig:gear pump}
\end{figure}

To demonstrate that the robot fish can also be powered with different power sources, a miniaturized gear pump(as shown in \Cref{fig:gear pump}) is also developed and tested~\cite{katzschmann_cyclic_2016}. The chambers are filled with water and attached to the pump outputs. An Electronic Speed Controller\footnote{ROXXY BL-Control 908 with \SI{8}{A} output current} is used to power the Brushless Direct Current (BLDC) motor\footnote{Park 300 Brushless Outrunner Motor, 1380Kv} which drives the miniaturized gear pump. When actuated, the pump generates pressure difference by forcing fluid from one chamber to the other periodically. The resulting pressure difference, with periodically reverting directions, is enough to create flapping motion in the tail. A video segment showing the flapping motion can be found in the supplementary material.

\section{Results}
\label{sec:results}

\begin{figure}[htbp]
            \centering
            \includegraphics[width=\columnwidth]{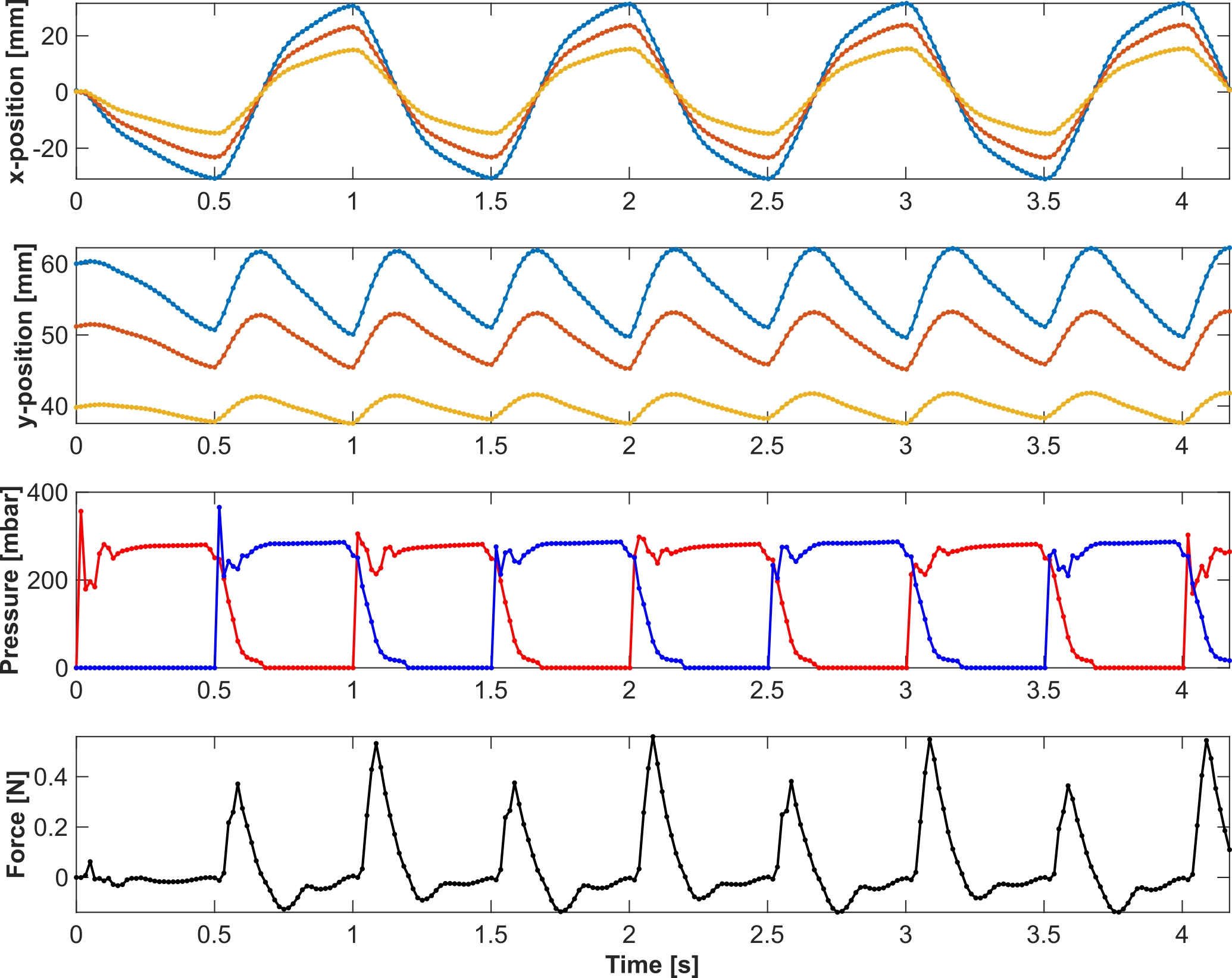}
            \caption{One of the obtained data plots, The experiment was done with Nemo with a pressure input of \SI{300}{\milli\bar} at \SI{1}{\Hz}. Here, $x$ is the left/right position and $y$ is the posterior position. Different colors corresponds to multiple tracked-dots at different locations along the tail.The blue and red colored lines in the pressure plot corresponds to the pressure measurements in the left and right chambers.}
            \label{fig:exp_data}
\end{figure}

The measured data from the experiments are processed to generate the final data set for the experiments. Black dots painted on the fish back are tracked in the video footage, their 2-dimensional positions in the horizontal plane is then corrected with the grid pad's reference and the camera's parameters. After identifying the video frame in which the LED lit up and aligning the position data, the time-synced records of pressure input, tail deformation, and thrust force can be obtained.
A data plot taken with Nemo\footnote{DS10 silicone with 9 chambers} is shown in \Cref{fig:exp_data} as an example.

\section{Discussion and Conclusion}
\label{sec:conclusion}
\subsection{Discussion} 

A few observations can be made upon examination of the captured data shown in \Cref{fig:exp_data}.
It is shown that the tail actuator undergoes a periodic flapping motion with a frequency of \SI{1}{\Hz} when actuated with a \SI{1}{\Hz} pressure signal. The periodic flapping motion of the tail then generates a thrust force with peaks occurring at \SI{2}{\Hz}. Since the frequency of the flapping motion is defined to be based on full flapping cycles that consist of two swings in total (one to the left and one to the right lateral direction), the difference in doubling frequency between the flapping motion and the thrust force output matches as expected. 

\begin{figure}[htbp]
        \centering
        \includegraphics[width=\columnwidth]{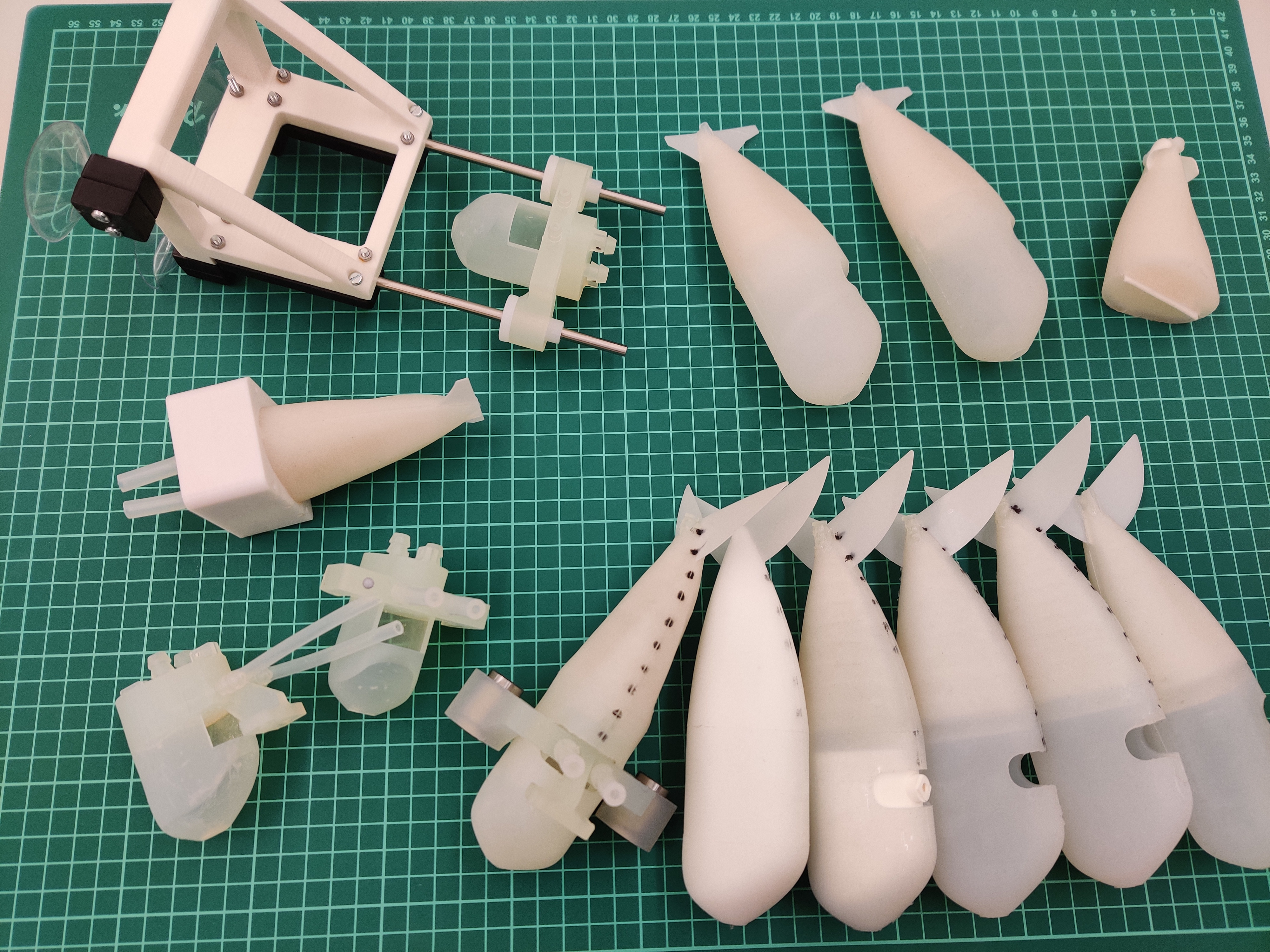}
        \caption{Some of the robots and test setups developed in this project. Every robotic fish / actuator has a slight difference in terms of design and material which further demonstrates the flexibility of the proposed development pipeline. The stand-alone tail actuators are used to verify the design and fabrication process in early stages. An early prototype of the test setup, where nylon bushings are used instead of linear bearings, is shown in the top-left corner. Apart from the 5 robotic fish described in \cref{sec:evaluation}, another prototype is constructed to explore the possibility of achieving neutral buoyancy. A mixture of silicone elastomer and glass bubbles(3M Glass Bubbles K1 with 1:100 mixing ratio) is used to construct a robotic fish with a solid body}
        \label{fig:result_picture}
\end{figure}

Similar observations can be made from the remainder of the experimental data with different robot configurations and control inputs. The robotic fish can also be powered with different actuation methods as described in \Cref{sec:evaluation}. In the making of robots with different designs, only a minimal set of fabrication tools and components are changed. This means that a bigger range of designs with various body shape and different actuation methods can also be fabricated with ease. For example, the wax cores can be further modified to create two single chambers in the tail section where actuators such as HASELs can be installed. Robots with a longer body section can also be realized with a simple swap of the body section molds to form a larger body cavity space where electronics for better sensing, control, and power storage can be installed. A picture of some the robotic fish, actuators, and test setup is shown in \Cref{fig:result_picture}.
We can thus conclude that the design, fabrication, and evaluation pipeline proposed in this work can be successfully applied to rapidly verify different robotic fish designs thanks to its modularity and flexibility. The simple and fast fabrication process will enable the mass production of robotic fish to be used in studying the swarming and schooling behavior of natural underwater life.
The proposed work can also serve as a standardized development pipeline for future projects involving robotic fish. This work will equip the soft robotics community with a powerful tool to improve research speed and efficiency on projects related to modeling, simulation, and design optimization. Such improvements can further lead to a better understanding of underwater locomotion and open up a world of possibilities with efficient bio-inspired robotic designs. The newly constructed robots can serve important roles in underwater explorations~\cite{katzschmann_exploration_2018}, search and rescue missions, and link a strong connection between engineering and studies of nature. With improved understanding of evolution, humans, the most delicate creation of nature's most powerful tool, might even be able to learn more about themselves someday.

\subsection{Future Work}
The presented work left open various aspects in which a future study can certainly dive deeper into. Those aspects are:
\begin{enumerate}
    \item Since current designs still require minimal manual adjustment between iterations, a parametric design can be created to allow for automatic design changes and further improve iteration efficiency.  
    \item The design can be of benefit for a bigger community if the design files would be fully documented and then open-sourced for the general public.
    \item The two tested actuation methods belong to the type of fluidic actuation~\cite{katzschmann_cyclic_2016}. More actuation methods such as HASELs~\cite{acome_hydraulically_2018}, DEAs~\cite{ohalloran_review_2008}, and bio-hybrid muscles~\cite{carlsen_bio-hybrid_2014} will be an exciting research direction to explore. 
    \item Current designs of the robotic fish are still incapable of un-tethered operation. Further developments for a fully integrated robot can allow the robot to be used in a wider range of missions.
\end{enumerate}








\section*{ACKNOWLEDGMENT}
We are grateful for the support by John Z. Zhang in the data processing and visualization process, and
the guidance and suggestions from Mr. Thomas Buchner during the project development period.
We are also grateful for the guidance and suggestions from Stephan-Daniel Gravert and Barnabas Gavin Cangan during the preparation of the supplementary materials for this work.
We are grateful for the advice and feedback from Mike Yan Michelis and Sebastian Pinzello during the manuscript writing.


\bibliographystyle{IEEEtran}
\bibliography{09_references}

\end{document}